Article
# Making Sense of Complex Sensor Data Streams

**Rongrong Liu, Birgitta Dresp-Langley**

1. ICube Lab, Robotics Department, Strasbourg University, 67085 Strasbourg, France
2. ICube Lab, UMR 7357 Centre National de la Recherche Scientifique CNRS, Strasbourg University, 67085 Strasbourg, France
* Correspondence: rongrong.liu@unistra.fr (R.L.) or birgitta.dresp@unistra.fr (B.D.-L.)

**Abstract:** This concept paper draws from our previous research on individual grip force data collected from biosensors placed on specific anatomical locations in the dominant and non-dominant hand of operators performing a robot-assisted precision grip task for minimally invasive endoscopic surgery. The specificity of the robotic system on the one hand, and that of the 2D image-guided task performed in a real-world 3D space on the other, constrain the individual hand and finger movements during task performance in a unique way. Our previous work showed task-specific characteristics of operator expertise in terms of specific grip force profiles, which we were able to detect in thousands of highly variable individual data. This concept paper is focused on two complementary data analysis strategies that allow achieving such a goal. In contrast with other sensor data analysis strategies aimed at minimizing variance in the data, it is necessary to decipher the meaning of intra- and inter-individual variance in the sensor data on the basis of appropriate statistical analyses, as shown in the first part of this paper. Then, it is explained how the computation of individual spatio-temporal grip force profiles allows detecting expertise-specific differences between individual users. It is concluded that both analytic strategies are complementary and enable drawing meaning from thousands of biosensor data reflecting human performance measures while fully taking into account their considerable inter- and intra-individual variability.

## 1. Introduction

Wireless technology provides the ability to communicate or control over distances without requiring of wires or cables, or any other electrical conductors, but using electro-magnetic waves, which were first conclusively proved to exist by the German physicist Heinrich Hertz. After continuous efforts over a century, many types of wireless systems have emerged, such as Infrared, Bluetooth, Wireless-Fidelity (WIFI), Radio Frequency Identification (RFID), Global Positioning System (GPS), etc. Presently, wireless technology plays a key role throughout the world, and the applications could be found for communications in cities [1], public buildings [2], individual houses [3], cars [4], people [5], animals [6].

Along with the advance of wireless technology, and other technique innovation in sensor design, electronic and power management, wireless wearable biosensor sys-tems [7,8] are currently developing rapidly, which can convert a biological response into electrical measurements using electronic circuits, and allow transmiting the detected infor-mation remotely without using cables or wires, to a data acquisition platform [9]. Unlike conventional approaches, these devices enable convenient, continuous, unobtrusive and real-time monitoring and analysis [10] of signals, including chemical signals such as gas and biomolecules, thermals signals such as fever and hypothermia, electrophysiological signals such as brainwave and cardiac activities, and physical signals such as pressure, motion and, as will be shown in this paper, individual grip force data.

Individual grip force distributions necessary to complete functional tasks exhibit functional redundancies and depend on the type of task [11]. For example, when a cylin-drical object needs to be manipulated, the middle finger will play an important role in the generation of gross force contribution, especially when the task consists of lifting a heavy object. In spherical grasp patterns, the contribution of the ring and small fingers is important to the total grip force. Individual finger forces have also been found to differ by gender [12] and age [13], indicating changes in the processing of fine motor control tasks with increasing age, presumably caused by difficulties of late middle-aged adults to produce any required amount of force rapidly.

This paper here is focused on data analysis strategies for the specific case of a precision grip task for the control of a robot-assisted surgical platform. The specific characteristics of the hard and software components of the robotic system are described in full detail in our previous publications [10,14–18]. The robotic surgical system [19] is a prototype, and data are currently available from one highly proficient expert user (with a total number of about 116 thousand of data available), one moderately trained user (153 thousand) and, for comparison, one complete novice user (171 thousand).

In a first part, we describe how issues relative to intra- and inter-individual variance in the data need to be dealt in the specific case of human precision grip force deployment. This involves particular statistical tools and metrics. Data variance in biosensor networks, such as the grip force sensor network in the human hand explored here in this Figure 1, is directly related to specific functional differences between locations where the sensors are placed (our own work), and needs to be fully taken into account for the interpretation of the data under the light of specific task constraints on the one hand, and biomechanical constraints directly related to hand and finger movements during grip force deployment on the other. This sheds a radically different light on analysis of variance by comparison with cases where variance in the sensor data needs to be eliminated or minimized [20].

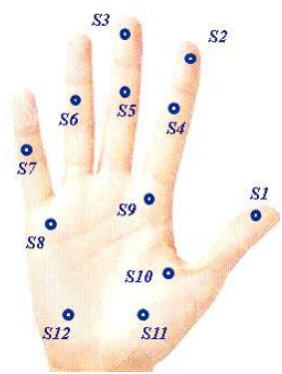

**Figure 1.** Sensor locations [17].

In a second part, we illustrate a method, which will be described in detail in the data analysis section, that permits detecting individual spatio-temporal profiles in thousands of variable grip force data by focusing on task specific finger locations. How can help under-stand expertise-specific differences in grip force deployment between a highly proficient operator, a trainee, and a complete novice, and how these anatomically specific and task relevant grip forces evolve with time and level of training, will be shown.

The remainder of the paper is organised as follows. In Section 2, the materials and methods are explained, including the experiment platform, sensor glove and experiment design. The results are presented and the statistical analysis on time data, force data and functionally representative sensors are investigated in Section 3. In Section 4, the discussion of this study and some ideas for future work are proposed.

**2. Materials and Methods**
2.1. Experimental Platform

The experiment has been conducted on a robotic endoscope system called STRAS, which stands for Single access and Transluminal Robotic Assistant for Surgeons [19,21], aiming to optimally assist surgeons in minimally invasive procedures. It is designed for bi-manual intervention, and there is an endoscopic camera attached at the distal side of the robotic endoscope and filming the users performing the task. During the manipulation of the STARS endoscope system, as shown in Figure 2, the user can hold two joystick handles to move and manipulate the endoscope tools in each hand while facing a monitor screen that displays the camera view.

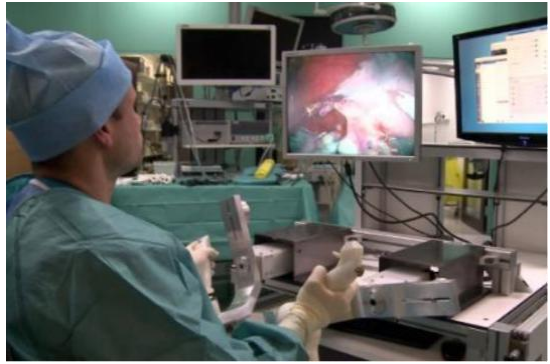

**Figure 2.** STRAS endoscope manipulation.

2.2. Sensor Glove Design

A pair of specific wearable wireless sensor gloves were developed [14,15], using inbuilt Force Sensitive Resistors (FSR). Each of the small (5–10 mm diameter) FSR was soldered to 10KW pull-down resistors to create a voltage divider. The voltage read by the analog input of the Arduino is given by Equation (1)

$$V_{out} = R_{PD} V_{3.3} / (R_{PD} + F_{FSR}) \tag{1}$$

where $R_{PD}$ the resistance of the pull down resistor, $R_{FSR}$ he FSR resistance, and $V_{3.3}$ is the 3.3 V supply voltage. FSR resistances can vary from 250 W when subjected to 20 Newton (N) to about 10 MW when no force is applied at all. The voltage varies monotonically between 0 and 3.22 Volt, as a function of the force applied, which is assumed uniform on the sensor surface. In the experiments here, forces applied did not exceed 10 N; voltages varied within the range of [0; 1500] mV. The relation between force and voltage is almost linear within this range. It was ensured that all sensors provided similar calibration curves.

These FSR have been first glued inside the glove, and then secured by sewing a circular piece of cloth around. For each glove, which is shown in Figure 3, twelve anatomically relevant FSR are employed to measure the grip force applied on certain locations on the fingers and in the palm, as illustrated in Figure 1.

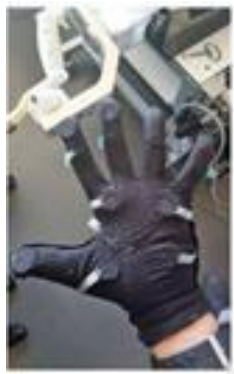

**Figure 3.** Glove surface.

The software to acquire grip force data includes two parts: one running on the Arduino Micro board embedded on the gloves, and the other running on the computer for data collection. Figure 4 shows this data acquisition process by taking Sensor 10 (S10) as an example. Powered by Li-Po battery, the Micro board provides a regulated voltage, acquires analog voltage output from each FSR sensor, which is merged with the time stamps and sensor identification. This data package is then sent to the computer wirelessly via Bluetooth with a frequency of 50 Hz, decoded by the computer software, and saved in an excel file.

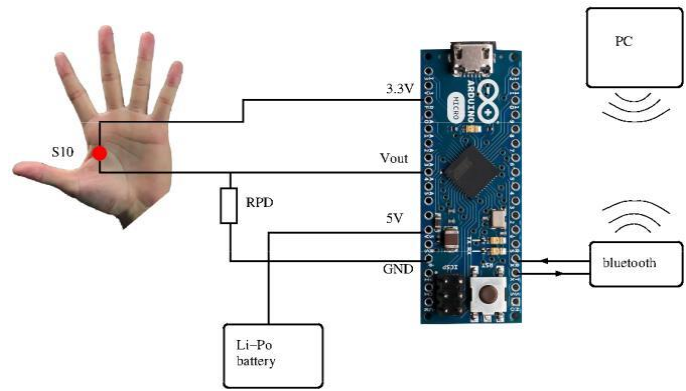

**Figure 4.** Diagram of data acquisition system [16].

2.3. Experimental Design

Equipped with the STRAS platform and the sensor gloves, a pick-and-drop image-guided robot-assisted task was designed for this individual grip force study. The experiment consists of manipulating the STRAS endoscope system with one hand while wearing the sensor glove. The precision grip task is entirely image-guided [10,14–18] and executed by a highly proficient expert user, a moderately trained user, and a complete novice. The expert is left handed while the trained user and the novice are right handed, as illustrated in Table 1. The left and right interfaces are identical, and this pick-and-drop task has been realized with each hand here for ten successive sessions.

The specificity of the precision grip task in this case here may, on the one hand, be described by the fact that the degrees of freedom for hand and finger movements are unusually constrained. To operate the cylindrical handles in order to direct the surgical tools, and to open and close the grippers at the tool-ends during the surgical task, only sideways and forward/backward movements of the cylindrical handles are possible for manipulating the tool movements in all directions in the three-dimensional workspace. This is radically different from gripping cylindrical objects to lift and/or move them around directly and freely in all possible directions. On the other hand, any surgical task executed with the system, including the simulator task used for training and for which data were collected and exploited in this paper here, imposes further constraints in terms of specific task steps. These need to be executed one after the other in a precise order with the least of effort, the greatest precision, and as swiftly as possible.

Essentially, this experiment consists of four critical steps. Figure 5 shows a snapshot for these four steps from the video sequence captured by the endoscopic camera. As also described in Table 2, to accomplish this four-step precision grip task, the user first activates and moves the distal tool toward the object by manipulating the handles effectively. During this step, movement along the depth in 2D image plane is required. When the tool arrives the object location, the grippers will be opened to grab and lift the object firmly. In the third step, the user has to move the tool with object until the target position, during which only lateral movement is needed. In the last step, the user has to open the grippers and drop the object [22–26].

Since the task here is entirely image-guided, the first task step here is the most difficult, especially for the novice, as it requires moving the tool-tips ahead in depth along an invisible z-axis in the 2D image plane. Even experts still have problems in adjusting to this problem, which can be quite challenging depending on the type of camera and imaging system available to the surgeon on a specific platform. Here in our case, to accomplish the task steps in optimal task time, and with maximal precision (i.e., no tool trajectory corrections, no incidents) requires not only being familiar with the 2D image projection of the three-dimensional task workspace, but also the skillful manipulation of the task-relevant (left or right) handle of the robotic system, as pointed out in Section 2 of our previous work [10,14–18].

**Table 1.** User Information

| User | Dominant Hand |
|---|---|
| Proficient expert | Left |
| Intermediate user | Right |
| Complete novice | Right |

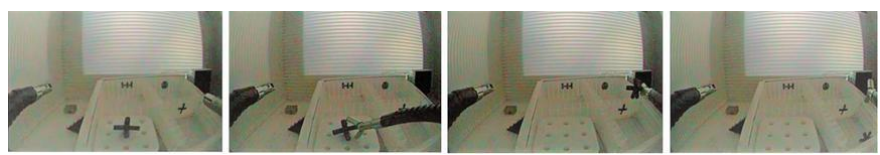

**Figure 5.** Snapshot views of the four successive steps [15].

**Table 2.** Four-step pick-and-drop task.

| Step | Description |
|---|---|
| 1 | Activate and move tool towards object location |
| 2 | Open and close grippers to grasp and lift object |
| 3 | Move tool with object to target location |
| 4 | Open grippers to drop object in box |

## 3. Results and Analysis

As mentioned in the previous section, this pick-and-drop task was realized with each hand of three users for ten successive sessions, which means the same task has been repeated for sixty times, a total of 60 time results have been recorded, together with 440,412 grip force signals having been collected from the twelve sensor locations on the dominant and non-dominant hands of three users in ten successive task sessions, corresponding to a total of 36701 grip force signals per sensor, as shown in Table 3.

The distinct levels of expertise are consistently reflected by performance task parameters such as average task session times, or the number of task incidents in terms of object drops, misses, and tool-trajectory adjustments during individual performance across task sessions. Left and right system interfaces are identical, and the same task is realized with either hand. The individual grip forces are centrally controlled in the human brain, and aimed at optimizing human motor performance and control [27–31].

**Table 3.** Number of grip force signals for each sensor.

| User | Dominant | Non-Dominant |
|---|---|---|
| Expert | 4442 | 5244 |
| Intermediate user | 5974 | 6764 |
| Novice | 7780 | 6497 |

3.1. Time Results and Analysis

As discussed above, there are three levels of user factor and the two levels of handness factor. A two-way ANalysis Of VAriance (ANOVA) has been conducted to access the effects of these two factors on the task execution time. As shown in Table 4, there is an extremely significant difference among three users, while there is no significant difference between dominant and non-dominant hands. Moreover, the last row illustrates a significant interaction between the user and handness factors, which means here that the execution times of different users depend on which hand they are using to accomplish the precision task.

**Table 4.** Results from two-way ANOVA on time data as a function of user and handness.

| Source of Variation | Degree of Freedom | F | P |
|---|---|---|---|
| User | 2 | 15.65 | <0.001 |
| Handness | 1 | 0.09 | Not significant |
| User Handness | 2 | 4.13 | <0.05 |

Time results across sessions are given in terms of means and their Standard Errors of Mean (SEM) in Table 5. Since the differences between means are difficult to grasp from looking at the tables, we also represent the time results graphically in Figure 6, where execution times by dominant and non-dominant hands are shown separately on top, and total amount of times of both hands for each session are illustrated at bottom, together with total execution times for each hand.

**Table 5.** Mean and SEM of task execution times across sessions

| User | Handness | Mean (s) | SEM (s) |
|---|---|---|---|
| Expert | Dominant | 8.88 | 0.36 |
|  | Non-dominant | 10.49 | 0.49 |
| Intermediate | Dominant | 11.95 | 0.49 |
|  | Non-dominant | 13.53 | 0.66 |
| Novice | Dominant | 15.56 | 1.60 |
|  | Non-dominant | 12.99 | 0.75 |

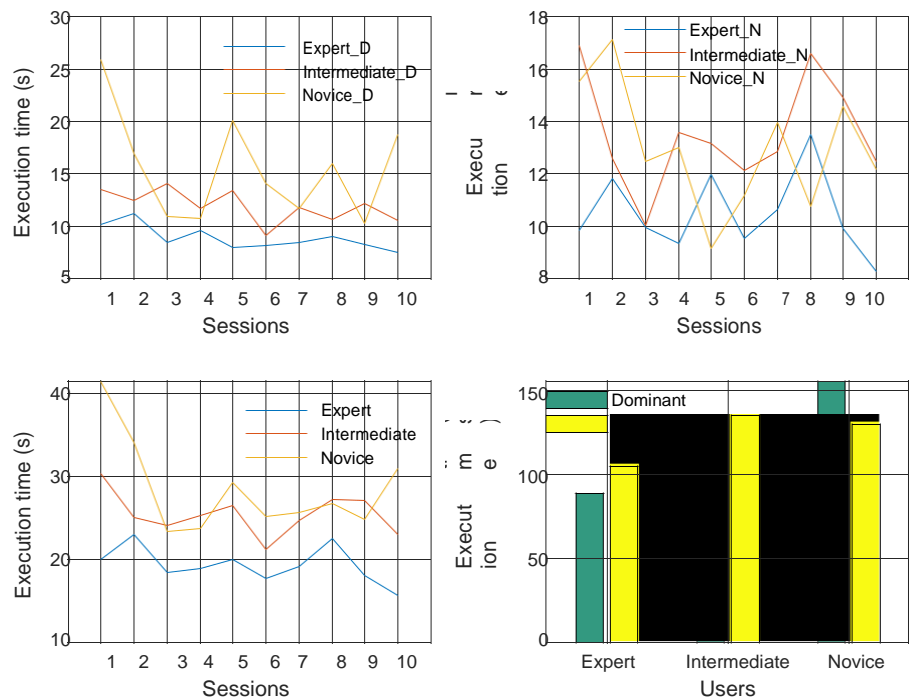

**Figure 6.** Execution time for three users with dominant and non-dominant hands among ten sessions.

Based on Table 5 and Figure 6, it is clear that the expert and the dominant hand of the intermediate user have a better performance based on time results, with smaller means and variations. Moreover, the novice user is slower and less stable with his dominant hand, which indicates that in this kind of extremely constrained environment of precision task, the novice user was more hesitant in manipulating with his dominant hand.

3.2. Force Results and Analysis

The locally grip forces deployed at each sensor location vary depending on how skilled a user has become in performing the robot-assisted simulator task accurately. In our work, grip force data were collected from three users with distinctly different levels of task expertise. As mentioned at the beginning of this section, there are much more force data than time data. To start with, total forces applied on each sensor during ten sessions by each hand are calculated from the original raw data, and given in Table 6 to offer a general description analysis. According to this table, different force strategies have been applied for different users and hands. Novice applied a significant higher total force compared with other two users, especially with the dominant hand. Moreover, some sensors have not been used during certain maneuvers. For example, the expert did not activate S1, S8 and S11 with his dominant hand, while S2 and S3 were not used with his non-dominant hand.

**Table 6.** Total force across sessions for each sensor (V).

| Sensor | Expert_D | Expert_N | Interm_D | Interm_N | Novice_D | Novice_N |
|---|---|---|---|---|---|---|
| 1 | 0 | 3.40 | 0 | 0.00 | 0 | 0 |
| 2 | 6.23 | 0 | 9.16 | 58.67 | 193.23 | 447.29 |
| 3 | 10.96 | 0 | 0 | 0 | 5328.26 | 0 |
| 4 | 9.03 | 46.90 | 37.50 | 0.60 | 6.07 | 0 |
| 5 | 437.13 | 1811.11 | 2901.79 | 283.75 | 5946.81 | 1926.19 |
| 6 | 2009.06 | 1895.47 | 3327.50 | 3520.78 | 3915.37 | 6910.31 |
| 7 | 2607.76 | 115.71 | 60.63 | 3638.08 | 664.06 | 3420.98 |
| 8 | 0 | 487.50 | 1064.15 | 38.27 | 5022.85 | 1512.46 |
| 9 | 2.50 | 534.02 | 786.74 | 0 | 8838.14 | 0.66 |
| 10 | 2106.27 | 900.44 | 489.66 | 499.17 | 5062.52 | 3246.43 |
| 11 | 0 | 3966.70 | 0 | 0 | 6842.42 | 0 |
| 12 | 5.15 | 2242.13 | 1593.11 | 0 | 6585.59 | 2.71 |
| Total | 7194.10 | 12,003.39 | 10,270.23 | 8039.33 | 48,405.31 | 17,467.03 |

In the case of force results, there are three levels of user factor, two levels of handness factor and twelve levels of sensor factor. A three-way ANOVA has been conducted to access the effects of these three factors on the force results. As shown in Table 7, there is an extremely significant difference of force employed among three users, between two hands and among twelve sensors, and an extremely significant interaction exists between each two factors. The corresponding means and SEM are provided in Table 8. Moreover, the average forces in different categories are also visualized in Figure 7.

**Table 7.** Force results from three-way ANOVA on force data as a function of user, handness and sensor.

| Source of Variation | Degree of Freedom | F | P |
|---|---|---|---|
| User | 2 | 107.72 | <0.001 |
| Handness | 1 | 38.27 | <0.001 |
| Sensor | 11 | 41.88 | <0.001 |
| User  Handness | 2 | 47.83 | <0.001 |
| User  Sensor | 22 | 7.41 | <0.001 |
| Handness  Sensor | 11 | 9.85 | <0.001 |

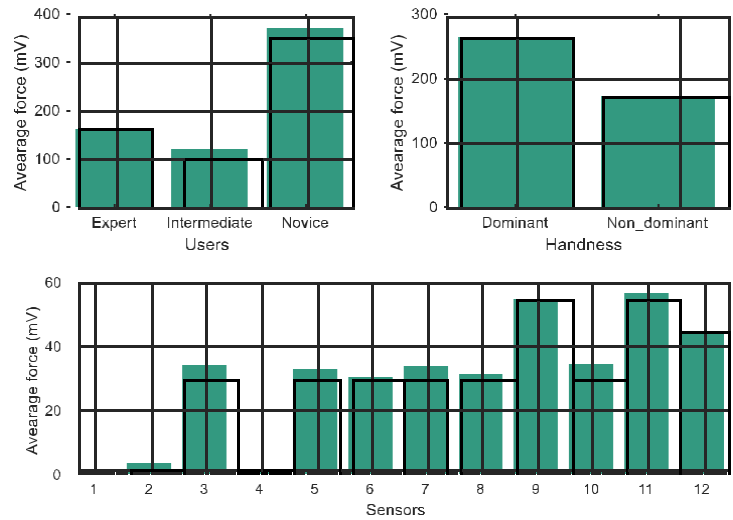

**Figure 7.** Force results for three users with dominant and non-dominant hands among ten sessions

**Table 8.** Force Mean and SEM for all sensors

| Factor | Level | Mean (mV) | SEM (mV) |
|---|---|---|---|
| User | Expert | 161.96 | 15.85 |
|  | Intermediate | 120.29 | 13.10 |
|  | Novice | 370.55 | 25.49 |
| Handness | Dominant | 263.74 | 17.93 |
|  | Non-dominant | 171.46 | 14.48 |
| Sensor | S1 | 0.12 | 0.07 |
|  | S2 | 17.60 | 3.60 |
|  | S3 | 115.70 | 34.08 |
|  | S4 | 3.05 | 0.54 |
|  | S5 | 337.63 | 32.69 |
|  | S6 | 575.04 | 30.21 |
|  | S7 | 294.61 | 33.76 |
|  | S8 | 193.40 | 31.19 |
|  | S9 | 223.82 | 54.57 |
|  | S10 | 320.96 | 34.51 |
|  | S11 | 267.99 | 56.54 |
|  | S12 | 261.28 | 44.41 |

3.3. Functionally Representative Sensors Analysis

In this part, the grip force data from three specific sensors, namely S5, S6, and S7, which are explained and shown in Figure 8 will be further analyzed, as they are functionally representative according to earlier studies [30,31]. Detailedly speaking, S5, which locates on the middle phalanx of the middle finger, mostly contributes to gross force deployment, such as lifting heavy objects, but useless for precision tasks. On the contrary, S7 is on the middle phalanx of the pinky finger and critically important in fine grip force control manipulation studied here. In addition, the last sensor S6, locating on the middle phalanx of the ring finger, is among the least important in grip force control across a variety of tasks.

| Sensor | Finger | Role in grip force control |
|--------|--------|---------------------------|
| S5 | Middle | Gross grip force deployment |
| S6 | Ring | No meaningful role in grip force control |
| S7 | Pinky | Precision grip force control |

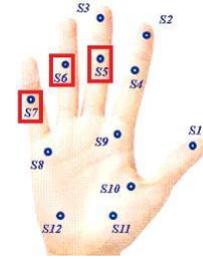

**Figure 8.** Three functionally representative sensors [17].

Based on the former analysis in this section, the most distinct performance happened between the dominant hands of expert and novice. Therefore, a two-way ANOVA has been conducted on the raw grip force data on these two hands of their first and last task sessions, and illustrated in Table 9. Statistical comparison reveals significant interaction between the two-level user and session factors for all three sensors considered here. To deliver a more direct comparison, individual spatio-temporal grip force profiles have been plotted in terms of average peak amplitude foe every one hundred signals for sensors S5, S6 and S7 in Figure 9, together with relative durations of each of the four task steps using the colored lines.

**Table 9.** Two-way ANOVA on three sensors on dominant hand of expert and novice for the first and last session (mV).

| Sensor | Session | Expert (Mean/SEM) | Novice (Mean/SEM) | Interaction |
|--------|---------|-------------------|-------------------|-------------|
| S5 | First | 240.37 /4.56 | 790.00 /3.02 | $F(1,3120) = 169.39$; |
|    | Last  | 48.32 /0.36  | 691.72 /2.19 | $p < 0.001$ |
| S6 | First | 575.63 /4.51 | 504.12 /2.42 | $F(1,3120) = 394.24$; |
|    | Last  | 473.98 /5.17 | 540.30 /2.23 | $p < 0.001$ |
| S7 | First | 594.02 /3.41 | 110.82 /0.75 | $F(1,3120) = 260.72$; |
|    | Last  | 608.51 /2.38 | 72.90 /0.61  | $p < 0.001$ |

One major difference between skilled or proficient operators and beginners concerns proportional gross grip force deployed by the middle finger. Novice operators deploy way too much unnecessary, task-irrelevant gross grip force, while the expert has learnt to skillfully minimize those [18]. On the other hand, precision grip forces, which are mostly deployed by the small finger and are particularly important in surgical tasks are generally insufficiently deployed by novices [18]. The ring finger plays no major role in grip force control, and the differences between ring finger grip force profiles of novices and experts can be expected to be minimal.

The spatio-temporal profile analysis here shows that the novice takes more than twice as long than the expert to accomplish the task, but at the end he scores a 30% time gain, indicating a considerable temporal training effect. This effect concerns mostly the first critical step in Figure 9 of the pick-and-drop task and becomes clear only under the light of the specific analysis provided here. The functional interpretation of this effect relates to the specificity of the tool-movement away from the body in the surgeon's peri-personal space required by this first task step. In the specific image-guided task-user system, grip forces and hand movements are constrained by the limited degrees of freedom of the robotic system on the one hand, and the perceptual recovery of physically missing depth information [18], here along a virtual z-axis in the 2D image plane, is necessary. Such difficulty results in longer task times and imprecise tool-movements, as shown in our previous work [10,14–18].

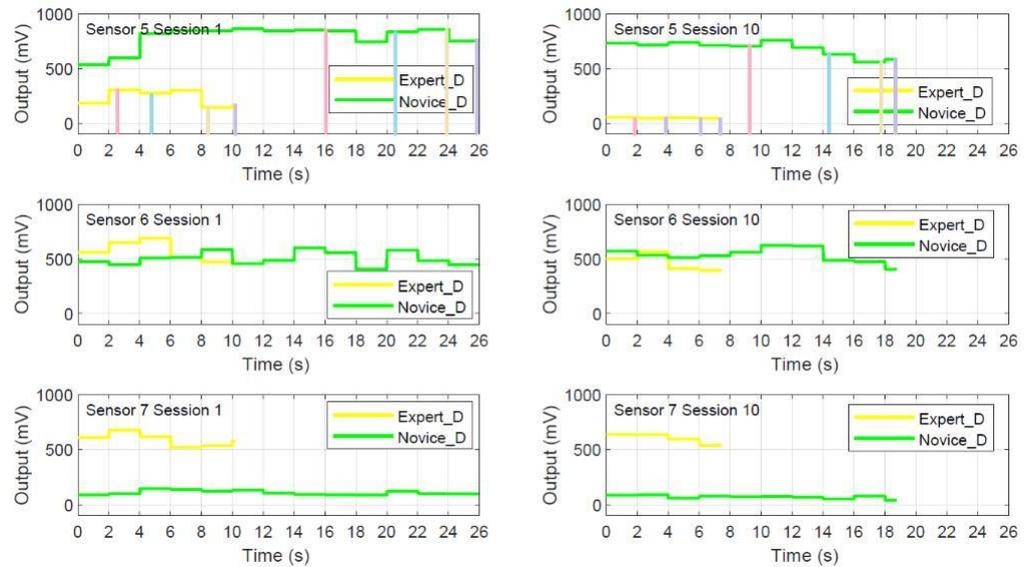

**Figure 9.** Average peak of three sensors on dominant hand of expert and novice for the first and last session [18]

## 4. Discussion

In this paper, we conducted spatio-temporal statistical analysis on the grip force data collected from three different individuals wearing wireless wearable sensor gloves while manipulating on an endoscope platform to accomplish a four-step pick-and-drop precision task. Following our results, as we could see in last sections, different users apply significant different grip force strategy based on their level of training and task expertise. As grip force monitoring can be run in real time during task execution, which could generate useful information for research on clinical decision making. For example, it could help prevent risks in robot assisted surgery [32], where excessive grip force may cause tissue damage. In future, if we have the opportunity to acquire more data, especially from the novice, the further study of the training effect and the force pattern development could be conducted, which is hopefully to deliver insight to junior surgeons training [33], exoskeletons devel-oping [34], rehabilitation robot design [35] and tactile internet implementation [36–38]. Moreover, we are building a new glove system with more advanced grip sensors [39], which are smaller and more flexible. Additionally, there is a clear need for more research on dynamic grip force measures that takes into account the hand-and- wrist complex in dy-namic force measurements [40,41]. The hand-and-wrist complex is particularly important in laparoscopic and endoscopic surgical tasks such as the one in this study here. The angle of hand-wrist movements directly influences hand and finger grip forces as a function of the diameter of the tool being grasped and manipulated. The grip forces then also depend on hand size [42]. Further insights into these functional relationships should be useful for the design of handles that require gripping in specific directions, to reduce the effort needed and to minimize surgeons' fatigue and exertion levels.

**Author Contributions:** Conceptualization, R.L. and B.D.-L.; methodology, R.L. and B.D.-L.; formal analysis, R.L. and B.D.-L.; investigation, R.L. and B.D.-L.; resources, R.L. and B.D.-L.; data curation, B.D.-L.; writing—original draft preparation, R.L. and B.D.-L.; writing—review and editing, B.D.-L.; visualization, R.L. and B.D.-L.; supervision, B.D.-L.; project administration, B.D.-L.; funding acquisition, B.D.-L. All authors have read and agreed to the published version of the manuscript.

**Funding:** This research work is part of a project funded by the University of Strasbourg's Initiative D'EXellence (IDEX).

**Data Availability Statement:** The raw data from our studies are accessible online by visiting our previously published work as cited.

**Acknowledgments:** The authors would like to thank Amine M. Falek, Florent Nageotte and Philippe Zanne, for hardware development and data acquisition. The support of the CNRS is also grate-fully acknowledged.

**Conflicts of Interest:** The authors declare no conflict of interest.

## Abbreviations

The following abbreviations are used in this manuscript:

| | |
|---|---|
| WIFI | WIreless-FIdelity |
| RFID | Radio Frequency Identification |
| GPS | Global Positioning System |
| STRAS | Single access and Transluminal Robotic Assistant for Surgeons |
| FSR | Force Sensitive Resistors |
| ANOVA | ANanalysis Of VAriance |
| SEM | Standard Errors of Mean |